\documentclass[nohyperref]{article}

\usepackage{microtype}
\usepackage{graphicx}
\usepackage{booktabs} 
\usepackage{caption}
\usepackage{subcaption}
\usepackage{enumitem}
\usepackage{hyperref}

\usepackage[accepted]{icml2022}

\usepackage{amsfonts}
\usepackage{amsmath}
\usepackage{tikz}
\usepackage{stmaryrd}
\usetikzlibrary{bayesnet}
\usepackage{graphicx} 

\usepackage{amsthm}
\usepackage{abraces}
\usepackage{amssymb}
\usepackage{dsfont}
\newcommand{\itemEq}[1]{%
        \begingroup%
        \setlength{\abovedisplayskip}{0pt}%
        \setlength{\belowdisplayskip}{0pt}%
        \parbox[c]{\linewidth}{\begin{flalign}#1&&\end{flalign}}%
        \endgroup}
 
\theoremstyle{plain}
\newtheorem{theorem}{Theorem}[section]

\theoremstyle{definition}
\newtheorem{definition}[theorem]{Definition}

\theoremstyle{remark}

\icmltitlerunning{Variational Inference for Infinitely Deep Neural Networks }

\begin{document}

\twocolumn[
\icmltitle{Variational Inference for Infinitely Deep Neural Networks }




\begin{icmlauthorlist}
\icmlauthor{Achille Nazaret}{to}
\icmlauthor{David Blei}{to,goo}
\end{icmlauthorlist}

\icmlaffiliation{to}{Department of Computer Science, Columbia University, New York, USA}
\icmlaffiliation{goo}{Department of Statistics, Columbia University, New York, USA}

\icmlcorrespondingauthor{Achille Nazaret}{achille.nazaret@columbia.edu}

\icmlkeywords{Machine Learning, ICML}

\vskip 0.3in
]



%
\printAffiliationsAndNotice{} 

\newcommand{\modelname}{unbounded-depth neural networks }
\newcommand{\Modelname}{Unbounded depth neural networks }
\newcommand{\modelnameshort}{UDN}

\newcommand{\R}{\mathbb{R}}
\newcommand{\N}{\mathbb{N}^{*}}
\newcommand{\X}{\mathbf{X}}
\newcommand{\x}{\textbf{x}}
\newcommand{\parens}[1]{\left( #1 \right)}
\newcommand{\bracket}[1]{\left[ #1 \right]}
\newcommand{\curly}[1]{\left\{#1\right\}}
\newcommand{\indep}{\perp \!\!\! \perp}
\newcommand{\bs}{\backslash}
\newcommand\bbrackets[1]{\left\llbracket#1\right\rrbracket}
\newcommand*{\E}[2][]{\mathbb{E}\ifx\\\left[#1\right]\\\else_{#1}\fi \left[#2\right]}
\newcommand{\kl}[2]{D_{\text{KL}}\parens{ #1 \| #2}}
\newcommand\ceil[1]{\left\lceil#1\right\rceil}
\newcommand{\bmat}[1]{\begin{bmatrix} #1 \end{bmatrix}}
\newcommand{\curlystack}[1]{
\left\{
\begin{array}{l}
 #1
\end{array}
\right.
}
\DeclareRobustCommand{\parhead}[1]{\textbf{#1}~}

\begin{abstract}

  We introduce the \textit{unbounded depth neural network} (UDN), an infinitely deep probabilistic model that adapts its complexity to the training data. The UDN contains an infinite sequence of hidden layers and places an unbounded prior on a truncation $\ell$, the layer from which it produces its data. Given a dataset of observations, the posterior UDN provides a conditional distribution of both the parameters of the infinite neural network and its truncation. We develop a novel variational inference algorithm to approximate this posterior, optimizing a distribution of the neural network weights and of the truncation depth $\ell$, and without any upper limit on $\ell$. To this end, the variational family has a special structure: it models neural network weights of arbitrary depth, and it dynamically creates or removes free variational parameters as its distribution of the truncation is optimized. (Unlike heuristic approaches to model search, it is solely through gradient-based optimization that this algorithm explores the space of truncations.) We study the UDN on real and synthetic data. We find that the UDN adapts its posterior depth to the dataset complexity; it outperforms standard neural networks of similar computational complexity; and it outperforms other approaches to infinite-depth neural networks.

\end{abstract}

\section{Introduction}
Deep neural networks have propelled research progress in many domains~\citep{goodfellow2016deep}.  However, selecting an appropriate complexity for the architecture of a deep neural network remains an important question: a \mbox{network} that is too shallow can hurt the predictive \mbox{performance}, and a network that is too deep can lead to overfitting or to \mbox{unnecessary} complexity.

In this paper, we introduce the \textit{unbounded depth neural network} (UDN), a deep probabilistic model whose size adapts to the data, and without an upper limit. With this model, a practitioner need not be concerned with explicitly selecting the complexity for her neural network.

A UDN involves an infinitely deep neural network and a latent truncation level $\ell$, which is drawn from a prior. The infinite neural network can be of arbitrary architecture and containinterleave different types of layers. For each datapoint, it generates an infinite sequence of hidden states $(h_k)_{k \geq 1}$ from the input $x$, and then uses the hidden state at the truncation $h_{\ell}$ to produce the response $y$. Given a dataset, the posterior UDN provides a conditional distribution over the neural network's weights and over the truncation depth. A UDN thus has access to the flexibility of an infinite neural network and, through its posterior, can select a distribution of truncations that best describes the data. The model is designed to ensure that $\ell$ has no upper limit.  


We approximate the posterior UDN with a novel method for variational inference.  The method uses a variational family that employs an infinite number of parameters to cover the whole posterior space. However, this family has a special property: Though it covers the infinite space of all possible truncations, each member provides support over a finite subset. With this family, and thanks to its special property, the variational objective can be efficiently calculated and optimized.  The result is a gradient-based algorithm that approximates the posterior UDN, dynamically exploring its infinite space of truncations and weights.


We study the UDN on real and synthetic data. We find that (i) on synthetic data, the UDN achieves higher accuracy than finite neural networks of similar architecture (ii) on real data, the UDN outperforms finite neural networks and other models of infinite neural networks (iii) for both types of data, the inference adapts the UDN posterior to the data complexity, by exploring distinct sets of truncations.

In summary, the contributions of this paper are as follows:
\begin{itemize}

\item We introduce the unbounded depth neural network: an infinitely deep neural network which can produce data from any of its hidden layers. In its posterior, it adapts its truncation to fit the observations.

    \item We propose a variational inference method with a novel variational family. It maintains a finite but evolving set of variational parameters to explore the unbounded posterior space of the UDN parameters.

    \item We empirically study the UDN on real and synthetic data.  It successfully adapts its complexity to the data at hand.  In predictive performance, it outperforms other finite and infinite  models.
\end{itemize}

This work contributes to research in architecture search, infinite neural networks, and unbounded variational families.  Section \ref{sec:relatedwork} provides a detailed discussion of related work.

\section{\Modelname}
\label{sec:model}
We begin by introducing the family of unbounded-depth neural networks. Let $\mathcal{D} = \curly{(x_i,y_i)}_{i=1}^n$ be a dataset of $n$ labeled observations. 

\parhead{Classical neural networks.} A classical neural network of depth $L$ chains $L$ successive functions $(f_\ell)_{1\leq \ell \leq L}$, eventually followed by an output function $o_L$.
$$
\begin{array}{cccccccccc}
  &          & f_1(.)      &          & f_2(.)      &          & f_3(.)      &          & \hdots & f_L(.)\\
  & \nearrow & \downarrow  & \nearrow & \downarrow  & \nearrow & \downarrow  & \nearrow &        & \downarrow\\
x &          & h_1         &          & h_2         &          & h_3         &          &        & h_L \\
  &          &             &          &             &          &             &          &        & \downarrow\\
  &          &             &          &             &          &             &          &        & o_L(.)\\
\end{array}
$$
Each $h_\ell$ is called a \textit{hidden state} and each $f_\ell$ is called a \textit{layer}.  Each layer is usually composed of a linear function, an activation function, and sometimes other differentiable transformations like \textit{batch-normalization} \cite{ioffe2015batch}. 
In deep architectures, $f_\ell$ can refer to a \textit{block} of layers, such as a succession of 3 convolutions in a Resnet \cite{he2016resnet} or a dense layer followed by attention in transformers \cite{vaswani2017attention}.

We fix a \textit{layer generator} $f$ which returns the layer $f_\ell$ for each integer $\ell$.  The layer generator can return layers of different shapes or types for different $\ell$ as long as two consecutive layers can be chained by composition. Similarly, we fix an \textit{output generator} $o_L$ which transforms the last hidden state $h_L$ into a parameter suitable for generating a response variable. We write $\theta_\ell$ for the parameters of $f_\ell$ and incorporate those of $o_L$ into $\theta_L$. With this notation, a finite neural network of depth $L$, generated by $(f,o)$, is written as $$\Omega_L = o_L \circ f_L \circ f_{L-1} \circ ... \circ f_2 \circ f_1$$ and has parameters $\parens{\theta_1, ..., \theta_L}$.

Finally, we fix a distribution $p$ parametrized by the output of the neural network $\Omega_L(x_i; \theta_{1:L})$, and use it to model the responses conditional on the input: \begin{align*}
    \forall i ,\quad  y_i | x_i, \theta_{1:L} &\sim p(y_i; \Omega_L(x_i; \theta_{1:L})).
\end{align*} Given a dataset $\{x_i, y_i\}_{i=1}^{N}$ and Gaussian priors on $\theta_{1:L}$, MAP estimation of the neural network parameters corresponds to classical methods for fitting a neural network with weight decay~\citep{neal1996priors}.  In this paradigm, the form of $p$ is related to the loss function \cite{murphy2012machine}.

This classical model requires that the layers be set in advance. Its flexibility and ability to capture the data depend crucially on the selection of $L$.  Too large and the model is too flexible, and can overfit.  Too small and the model is not flexible enough. It is appealing to consider a model that can adapt its depth to the data at hand.

\parhead{Unbounded depth neural networks.} We extend the finite construction above to formulate an unbounded depth neural network (UDN). We consider an infinite sequence of hidden states $(h_k)_{k \geq 1}$ generated by $(f_\ell)_{\ell \geq 1}$ and parametrized by an infinite set of weights $\theta \triangleq \parens{\theta_\ell}_{\ell \geq 1}$.

A challenge to conceptualizing an infinite-depth neural network is where to hang the observation. What we do is posit a possible output layer $o_\ell$ after each layer of hidden units.
$$
\begin{array}{cccccccccc}
  &          & f_1(.)      &          & f_2(.)      &          & f_3(.)      &          & \hdots \\
  & \nearrow & \downarrow  & \nearrow & \downarrow  & \nearrow & \downarrow  & \nearrow &        \\
x &          & h_1         &          & h_2         &          & h_3         &          & \hdots \\
  &          & \downarrow  &          & \downarrow  &          & \downarrow  &          &        \\
  &          & o_1(.)      &          & o_2(.)      &          & o_3(.)      &          & \hdots \\
\end{array}
$$
We then add an additional parameter, a truncation level $\ell$, to determine which $o_\ell$ will generate the response.

\begin{figure}
    \centering
    \includegraphics[]{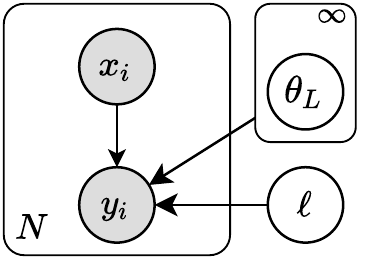}
    \caption{Graphical model for the unbounded depth neural network.}
    \label{fig:graphical_model}
\end{figure}

The complete UDN models the truncation level $\ell$ as an unobserved random variable with a prior $\mu$. Along with a prior $\rho$ on the weights, the UDN defines a generative model with an infinite-depth neural network: \begin{align*}
    \theta &\sim \rho(\theta) &&  \quad \triangleright \text{network weights} \\
    \ell & \sim \mu(\ell) && \quad \triangleright \text{truncation} \\
y_i | x_i, \theta, \ell &\sim p(y_i; \Omega_\ell(x_i ; \theta))  && \quad \triangleright \text{response} 
\end{align*}
This generative process is represented in figure \ref{fig:graphical_model}.  If the truncation prior $\mu$ puts a point mass at $L$ then the model is equivalent to the classical finite model of depth $L$. But with a general prior over all integers, the posterior UDN has access to all depths of neural networks.

The independence of the priors is important.  The model does not put a prior on $\ell$ and then samples the weights conditional on it. Rather, it first samples a complete infinite neural network, and then samples the finite truncation to produce its data. What this generative structure implies is that different truncations will share the same weights on their shared layers. As we will see in section \ref{sec:implementation}, this property leads to efficient calculations for approximate posterior inference.

\section{Variational Inference for the UDN}
\label{sec:inference}
Given a dataset $\mathcal{D} = \curly{(x_i,y_i)}_{i=1}^n$, the goal of
Bayesian inference is to compute the posterior UDN
$p(\theta, \ell \mid \mathcal{D})$. The exact posterior is
intractable, and so we appeal to variational
inference~\cite{jordan1999introduction, wainwright2008graphical,
  blei2017variational}.  In traditional variational inference, we
posit a family of approximate distributions over the latent variables
and then try to find the member of that family which is closest to the
exact posterior.

The unbounded neural network, however, presents a significant
challenge to this approach---the depth $\ell$ is unbounded and the
number of latent variables $\theta$ is infinite.  To overcome this
challenge, we will develop an unbounded variational family
$q(\theta, \ell)$ that is still amenable to variational
optimization. With the algorithm we develop, the ``search'' for a good
distribution of truncations is a natural consequence of gradient-based
optimization of the variational objective.

\newcommand{\Vfname}{Poly-bounded }
\newcommand{\vfname}{poly-bounded }
\subsection{Structure of the variational family}
We define a joint variational family that factorizes as
$q(\theta, \ell) = q(\ell) q(\theta | \ell)$, and note that the factor
for the neural network weights depends on the truncation. We introduce
the parameters $\lambda, \nu$ and detail the structure of the families
$\curly{q(\ell; \lambda)}$ and $\curly{q(\theta | \ell; \nu)}$.

\parhead{The unbounded variational family with connected and bounded members $q(\ell; \lambda)$.}
For a truly unbounded procedure, we require that the variational
family over $\ell$ should be able to explore the full space of
truncations $\N$. Simultaneously, since the procedure must run in
practice, each distribution $q(\ell; \lambda)$ should be tractable,
that is $\mathbb{E}_{q(\ell)}[g(\ell)]$ can be computed efficiently
for any $g$.

A sufficient condition for tractable expectations is that
$q(\ell; \lambda)$ has finite support; the expectation becomes the
finite sum $\sum_{i \in \text{support}(q)} q(i; \lambda) g(i)$.
However, to be able to explore the unbounded posterior space $\N$, the
variational family $\curly{q(\ell; \lambda)}$ itself cannot have
finite support. It should contain distributions covering all possible
truncations $\ell$. Moreover, it should be able to navigate
continuously between these distributions.

We articulate these conditions in the following definition:
\begin{definition}
  \label{def:varfam}
A variational family $\mathcal{Q} = \curly{q(\lambda)}$ over $\N$ is \textit{\mbox{unbounded} with connected and bounded members} if
\begin{enumerate}
  \itemsep-0.15cm 
    \item[\itemEq{\label{cond:bounded}}] $\forall q \in \mathcal{Q}, ~~ \text{support}(q) $ is bounded
    \item[\itemEq{\label{cond:mode}}] $\forall L \in \N, ~~ \exists q \in \mathcal{Q}, ~~L \in \text{argmax}(q)$ 
    \item[\itemEq{\label{cond:continuous}}] The parameter $\lambda$ is a continuous variable.
\end{enumerate}
\end{definition}
Echoing the discussion above, there are several consequences to this definition:
\begin{itemize}[leftmargin=*]
    \item By (\ref{cond:bounded}), each $q$ has a finite support. We write the maximal value $m(q) := \max \curly{\ell | q(\ell)>0}$.
    \item Thanks to (\ref{cond:mode}), the approximate posterior can place its main mass around any $\ell$. That is, $\mathcal{Q}$ covers the space of all possible truncations: $\bigcup_{q \in \mathcal{Q}} \text{support}(q) = \N$.
    \item Condition (\ref{cond:continuous}) ensures that $\mathcal{Q}$
      not only contains members with mass on any $\ell$, but it can
      continuously navigate between them. This condition is important
      for optimization.
\end{itemize}

\parhead{The nested family $q(\theta | \ell ; \nu)$.}
In the UDN model, conditional on $\ell$, the response $y$ depends only
on the first $\ell$ layers and not the subsequent ones. Thus the exact posterior $p(\theta | \ell)$ only contains
information from the data for $\theta_i$ up to $i \leq \ell$; the posterior of
the $\theta_i$ with $i > \ell$ must match the prior.

We mirror this structure in the variational approximation,
\begin{align}
  q(\theta | \ell; \nu) = q(\theta_{1:\ell}; \nu_{1:\ell})\prod_{k=\ell+1}^\infty p(\theta_k). \label{eqn:nested}
\end{align}
\citet{kurihara2007accelerated} also introduce a family with structure
as in (\ref{eqn:nested}), which they call a \textit{nested variational
  family}.

\parhead{The evidence lower bound.} The full variational distribution
combines the \textit{unbounded variational family} of Definition
\ref{def:varfam} with the \textit{nested family} of (\ref{eqn:nested}),
$q(\ell, \theta) = q(\ell; \lambda)q(\theta|\ell, \nu)$.  With this
distribution, we can now derive the optimization objective.

Variational inference seeks to minimize the KL divergence between the
variational posterior $q$ and the exact posterior $p$. This is
equivalent to maximizing the variational objective
\cite{bishop2006pattern}, which is commonly known as the Evidence
Lower BOund (ELBO). Because of the factored structure of the
variational family, we organize the terms of the ELBO with
iterated expectations,
\begin{align}
    \mathcal{L}(q)
    &= \mathbb{E}_{ q(\ell, \theta)}[\log p(Y,\ell, \theta | X) - \log q(\ell, \theta)] \\
    &= \mathbb{E}_{q(\ell)} \left[ \log
      \textstyle\frac{p(\ell)}{q(\ell)} + \mathbb{E}_{q(\theta \mid
      \ell)} \left[ \log \frac{p(\theta)}{q(\theta | \ell; \nu)}
      \right. \right. \nonumber  \\
    &\hspace*{1.3cm} \left. \left. + {\textstyle \sum_{i=1}^n }\log p(y_i \mid \ell, \theta, x_i) \textstyle\vphantom{\frac{p(\ell)}{q(\ell)}} \right] \right]. \label{eqn:finite-elbo}
\end{align}
Further, using the special structure of this variational family, the ELBO can
be simplified:
\begin{itemize}[leftmargin=*]
    \item The factor $q(\theta|\ell)$ satisfies the nested structure condition (\ref{eqn:nested}) so $ \frac{p(\theta)}{q(\theta | \ell; \nu)} =  \frac{p(\theta_{1:\ell})}{q(\theta_{1:\ell};\nu_{1:\ell})}$. This quantity only involves a finite number $\ell$ of parameters and variables even if the prior and posterior were initially over all the variables.
    \item The factor $q(\ell)$ satisfies (\ref{cond:bounded}).  The outer expectation of $\mathcal{L}(q)$ can be explicitly computed. 
\end{itemize}
With these two observations, we rewrite the ELBO:
\begin{align}
    \mathcal{L}(q)
    &=\sum_{\ell=1}^{m(q)} q(\ell; \lambda)\hspace{-0.1cm} \left[ \log \textstyle\frac{p(\ell)}{q(\ell;\lambda)} + \mathbb{E}_{q(\theta \mid \ell;\nu)} \hspace{-0.1cm} \left[\textstyle \sum\limits_{k=1}^\ell {\textstyle\log \frac{p(\theta_k)}{q(\theta_k ; \nu_k)}} \vphantom{\sum_{i=1}^\ell \log \frac{p(\theta_i)}{q(\theta_i | \ell; \nu_i)}} \right. \right. \nonumber  \\ 
    &\hspace*{2.2cm} \left. \left.  {  +  \textstyle \sum\limits_{i=1}^n }\log p(y_i; \Omega_\ell(x_i ; \theta)) \right] \right]. \label{eqn:final-elbo}
\end{align}
Notice this equation expresses the ELBO using only a finite set of
parameters $\lambda, \nu_{1:m(q(\lambda))}$, which we call \textit{active} parameters. Thus we can compute the
gradient of the ELBO with respect to $(\lambda, \nu_{1:\infty})$, since only the
coordinates corresponding to the active parameters $\lambda, \nu_{1:m(q)}$ can be nonzero.
This fact allows us to optimize the variational distribution.




\parhead{Dynamic variational inference.}
In variational inference we optimize the ELBO of equation
(\ref{eqn:final-elbo}). We use gradient methods to iteratively
update the variational parameters $(\lambda, \nu_{1:\infty})$. Equation
(\ref{eqn:final-elbo}) just showed how to take one efficient gradient step, by only updating the active parameters, those with nonzero gradients. From there, a succession of gradient updates becomes possible and still involves only a finite set of parameters. Indeed, even if the special property (\ref{cond:mode}) of the variational family guarantees that $q(\ell; \lambda)$ can place mass on any $\ell$, and by doing so, can activate any parameter $\nu_\ell$ during the optimization, successive updates of finitely many parameters will still only affect finitely many parameters.

For instance, the inference
can start with $m(q(\lambda)) = 5$ active layers, and increase to $m(q(\lambda)) = 6$
after an update to $\lambda$ that favors a deeper truncation. The next gradient update of $(\lambda, \nu)$ will then affect $\nu_{6}$, which was not activated earlier. At
any iteration, the ELBO involves a finite subset of the parameters, but this set of active parameters naturally grows or shrinks as needed to approach the exact posterior.

Because the subset of active variational parameters evolves during the optimization, we refer to the method of combining the \textit{unbounded variational family} of Definition \ref{def:varfam} with the \textit{nested family} of (\ref{eqn:nested}), as \textit{dynamic variational inference}. We detail in section \ref{sec:implementation} how to run
efficient computations when we do not know in advance which variational parameters will be activated during the optimization.

\subsection{One explicit choice of variational family and prior}
We end the inference section by proposing priors for $(\ell, \theta)$ and an explicit variational family satisfying the unbounded family conditions (\ref{cond:bounded}, \ref{cond:mode}, \ref{cond:continuous}) and the nested structure (\ref{eqn:nested}).

\parhead{Choice of prior.}
We use a standard Gaussian prior over all the weights $\theta$. We set a Poisson($\alpha$) prior for $\ell$. More precisely, we have $\ell - 1 \sim \text{Poisson}(\alpha)$ because $\ell >0$. The mean $\alpha$ is detailed in the experiment details in the appendix.

\parhead{Choice of family $q(\ell; \lambda)$.}
To obtain the unbounded family from definition \ref{def:varfam}, we adapt the Poisson family $\mathcal{P} = \curly{\text{Poisson}(\lambda) \mid \lambda > 0}$ by truncating each individual distribution $q(\ast; \lambda) = \text{Poisson}(\lambda)$ to its $\delta$-quantile. $$q^{{\delta}}(\ell ; \lambda) \propto q(\ell; \lambda)\mathds{1}[\ell \leq \delta\text{-quantile}(q(*; \lambda))]$$ This forms the Truncated Poisson family $\mathcal{TP}(\delta) = \curly{q^\delta \mid q \in \mathcal{P}}$ and the following holds:
\begin{theorem}
\label{th:tp}
For any $\delta \in [0.5, 1[$, $\mathcal{TP}(\delta)$ is unbounded with connected bounded members. 
For $\delta=0.95$, we have
\begin{itemize}
    \item \itemEq{\lambda - \ln 2 \leq m(q^{{0.95}}(\lambda)) \leq 1.3 \lambda + 5 \label{tp:upper}}
    \item \itemEq{\forall n\in \N, ~~ n \in \textnormal{argmax}(q^{0.95}(n+0.5)) \label{tp:argmax}}
    \item \itemEq{\lambda > 0 \text{ ~is a continuous parameter.}\label{tp:continuous}}
\end{itemize}
\end{theorem}

Inequalities (\ref{tp:upper}) are shown in appendix using Poisson tail bounds from \citet{short2013improved} and bounds on the Poisson median from \citet{choi1994medians}.
It shows that $q^{{0.95}}(\lambda)$ satisfies the bounded support condition (\ref{cond:bounded}) with a support growing linearly in $\lambda$. The result (\ref{tp:argmax}) offers explicit distributions in $\mathcal{TP}(0.95)$ that satisfy the unbounded family condition (\ref{cond:mode}). Finally, (\ref{tp:continuous}) ensures the continuity condition (\ref{cond:continuous}). During inference, $\delta$ is set to $0.95$ and $\lambda$ is a variational parameter.

\parhead{Choice of family $q(\theta \mid \ell; \nu)$.} 
The variational posterior on $\theta$ controls the neural network
weights. In the nested structure (\ref{eqn:nested}), we model
$q(\theta_{1:\ell}; \nu_{1:\ell})$ with a mean-field
Gaussian.\footnote{Some literature uses a point mass for the
  variational distribution of global parameters like neural network
  weights \cite{kingma2013auto, lopez2019joint} Here, however, this
  choice would cause problems because it creates a mixed
  discrete-continuous family that leads to discontinuities in the objective.}. With the prior defined earlier, $q(\theta | \ell; \nu)$ becomes   $$q(\theta | \ell; \nu) = \mathcal{N}(\nu_{1:\ell}, I_{\ell})[\theta_{1:\ell}]\prod_{k=\ell+1}^\infty \mathcal{N}(0, 1)[\theta_k]. $$ 
We approximate $\mathbb{E}_{q(\theta | \ell)}[g(\theta_{1:\ell})]$ at the first order with $g(\mathbb{E}_{q(\theta | \ell)}[\theta_{1:\ell}]) = g(\nu_{1:\ell})$ for any $g$.




\parhead{Predictions.}
The UDN can predict the labels of future data, such as held-out data for testing. It uses the learned variational posterior  $q(\ell, \theta; \lambda, \nu)$, to approximate the predictive distribution of the label $y'$ of new data $x'$ as
\begin{align}
    p(\mathit{y_{\text{}}' \mid x_{\text{}}'},\mathcal{D}) &\approx \mathbb{E}_{q(\ell, \theta; \lambda, \nu)}[p(y'; \Omega_\ell(x'; \theta_{1:\ell}))] \nonumber\\
    &\approx  \sum_{\ell = 1}^{m(q)} q(\ell; \lambda) \cdot p(y_{\text{}}'; \Omega_\ell(x'; \theta_{1:\ell})). \label{eqn:predictive}
\end{align} 
The predictive distribution forms an ensemble of different truncations, that is discovered during the process of variational inference. This is related to \citet{antoran2020depth}.

\section{Efficient algorithmic implementation}
\label{sec:implementation}
We review the computational aspects of both the model and the associated dynamic variational inference. 

\parhead{Linear complexity in $m(q)$.}
Evaluating the ELBO (\ref{eqn:final-elbo}) or evaluating the predictive distribution (\ref{eqn:predictive}) requires to compute the \mbox{output} of $m(q)$ different neural networks, $\Omega_1$ to $\Omega_{m(q)}$. However, most of the computations can be shared \cite{antoran2020depth}. We calculate the hidden layers sequentially up to $h_{m(q)}$, as they are needed to compute $\Omega_{m(q)}(x)$. We then apply the output layer $o_\ell$ to each hidden layer $h_\ell$ and obtain the collection $\curly{\Omega_\ell(x)}_{\ell=1}^{m(q)}$. Hence, computing $\Omega_{m(q)}(x)$ alone or the whole collection $\curly{\Omega_\ell(x)}_{\ell=1}^{m(q)}$ has the same complexity in $m(q)$. 

\parhead{Lazy initialization of the variational parameters.}
To compute gradients of the ELBO (\ref{eqn:final-elbo}), we leverage modern libraries of the Python language for automatic differentiation. As discussed in section \ref{sec:inference}, the gradient of the ELBO only involves a finite set of active parameters, yet, this set can potentially reach any size during the optimization. Hence, the ELBO can depend on every possible variational parameters $(\lambda, \nu_{1:\infty})$, not all of which can be instantiated.

In libraries like \textit{Tensorflow 1.0} \cite{tensorflow2015-whitepaper}, the computational graph is defined and compiled in advance. This prevents the dynamic creation of variational parameters. In contrast, a library like PyTorch \cite{pytorch2019} uses a dynamic graph. With this capability, new parameters $\nu_\ell$ and layers $f_\ell$ can be created only when needed and the computational graph be updated accordingly. Before each ELBO evaluation, we compute the support of the current $q(\ell; \lambda)$ and adjust the variational parameters. The full {dynamic variational inference} procedure is presented in Algorithm \ref{alg:example}. 

\begin{algorithm}
   \caption{Dynamic variational inference for the  UDN}
   \label{alg:example}
\begin{algorithmic}
   \STATE {\bfseries Input:} data $X, Y$; architecture generators $f, o$;
   \STATE Initialize: $\lambda$
   \STATE $hidden\_layers, output\_layers = [], []$
   \FOR{$epoch=1$ {\bfseries to} $T$}
   \STATE Compute $m(q(\lambda))$
   \WHILE{$L:=|hidden\_layers| < m(q(\lambda))$} 
   \STATE Add new layer $f(L+1)$ to $hidden\_layers$ 
   \STATE Add new layer $o(L+1)$ to $output\_layers$ 
   \STATE Initialize $\nu_{L+1}$
   \ENDWHILE
   \STATE Compute $\mathcal{L}(q)$ in a single forward pass
   \STATE Compute gradients $\nabla_{\lambda, \nu_{1:m(q(\lambda))}} \mathcal{L}(q)$
   \STATE Update $\lambda, \nu_{1:m(q(\lambda))}$
   \ENDFOR
\end{algorithmic}
\end{algorithm}

\section{Related work}
\label{sec:relatedwork}
\parhead{Neural architecture search}
Selecting a neural network architecture is an important question. Several methods have been proposed to answer it.

Bayesian optimization \cite{bergstra2013making,mendoza2016towards} and reinforcement learning \cite{zoph2016neural} can tune the architecture as a hyperparameter. These algorithms propose successive architectures, which are then evaluated and compared. These methods decouple the architecture search and the model training, which increases the computational complexity. Reusing parameters across runs \cite{pham2018efficient, luo2018neural} can speed up the search but cannot avoid multiple trainings. 

\citet{dikov2019bayesian, ghosh2019model} jointly learn the network weights and the architecture as a single model. This is similar to what the UDN can do. However, the architecture they learn can only be reduced from an initial candidate by masking some of its parts. This relates to network pruning methods, which remove connections of small weight \cite{lecun1990optimal, hassibi1992second}. 

Closely related to our work is \citet{antoran2020depth}, which combines predictions from different depths of the same model into a deep ensemble \cite{lakshminarayanan2017deepensembles} to improve its uncertainty calibration. Here, a maximal depth is specified by the user, and only networks of smaller depths are used. Setting a very high value may lead to unnecessary computational complexity. In contrast, the UDN can grow during training, only if necessary.

\parhead{Unbounded variational inference and Bayesian nonparametrics.}
An important class of models with an infinite number of latent variables is the class of Bayesian nonparametric (BNP) models \cite{hjort2010bayesian}. BNP models adapt to the data complexity by growing the number of latent variables as necessary during posterior inference. 

In a BNP model, variational inference does not usually operate in the unbounded latent space. Instead, proposed methods truncate either the model itself or the posterior variational family to a finite complexity \cite{blei2006variational, kurihara2007collapsed, doshi2009variational, ranganath2018correlated, moran2021spike}. 

To relax this restriction, some methods propose split-merge heuristics to create or remove latent variables during training \cite{kurihara2007accelerated,zhai2013online,hughes2013memoized}. In contrast, the dynamic variational inference proposed in the paper is a variational approach in which the creation or removal of variational parameters comes naturally from the variational family structure and gradient-based optimization.

\parhead{Infinite neural networks.} An infinite neural network does not have a definitive definition. We can distinguish two groups of concepts: infinite width and infinite depth.

\citet{neal1996priors} shows how a single-layer Bayesian neural network with infinite width and a judicious choice of weight prior is equivalent to a Gaussian process. The neural tangent kernel \cite{jacot2018neural, neuraltangents2020} describes how such a layer evolves during training. \citet{de2018gaussian, lee2018deep} extend these infinite-width results in multiple layers of deep networks. However, in recent work, \citet{pleiss2021limitations} indicate that a large width in a deep model can be detrimental.

Implicit models take a different approach, to the depth. Deep equilibrium \cite{bai2019deepequilibrium, bai2020multiscale} is obtained by constraining an infinite depth neural network to repeat the same layer function until reaching a fixed point. Neural ODEs \cite{chen2018neural, xu2021infinitelydeep, luo2022constructing} define a model with continuous depth -- hence infinite -- by extending the residual connection of a ResNet into differential equations.

However, these approaches all constrain the weights of the infinite neural network. Some place specific priors on the weights for infinite-width while others set constraint equations on the weights for infinite-depth. The resulting functions can be difficult to relate to classical neural networks. In contrast, the UDN offers flexibility on how to construct the network, while remaining similar to a classic neural network. The practitioner can use arbitrary architectures in a UDN and inspect the posterior like an ensemble of finite networks.

\section{Experiments}
\label{sec:experiments}
We study the performances of the UDN on synthetic and real classification tasks. We measure its predictive accuracy on held-out data and analyze how the posterior inference explores the space of truncations. 

We find that:
\begin{itemize}[leftmargin=*]
    \item The dynamic variational inference effectively explores the space of truncations. The UDN adapts its depth to the complexity of the data, from a few layers to almost a hundred layers on image classification. 
    \item The UDN posterior predictive accuracy outperforms all the finite network models $\Omega_L$. It also outperforms ensembles of finite networks \cite{antoran2020depth} and implicit models \cite{dupont2019augmented, bai2020multiscale}.
\end{itemize}

The experiments are implemented in Python with the library \textit{PyTorch} \cite{pytorch2019}. To scale the inference to large datasets, we use stochastic optimization on the ELBO \cite{robbins1951stochastic, hoffman2013stochastic}. This requires unbiased gradients of the ELBO (\ref{eqn:final-elbo}), which we compute by subsampling a batch of $256$ observations $(x_i, y_i)$ at each iteration.
The code is available on GitHub\footnote{ \url{https://github.com/ANazaret/unbounded-depth-neural-networks}}. 

\subsection{Spiral classification with fully connected networks}
\begin{figure}
    \centering
    \includegraphics[width=\linewidth]{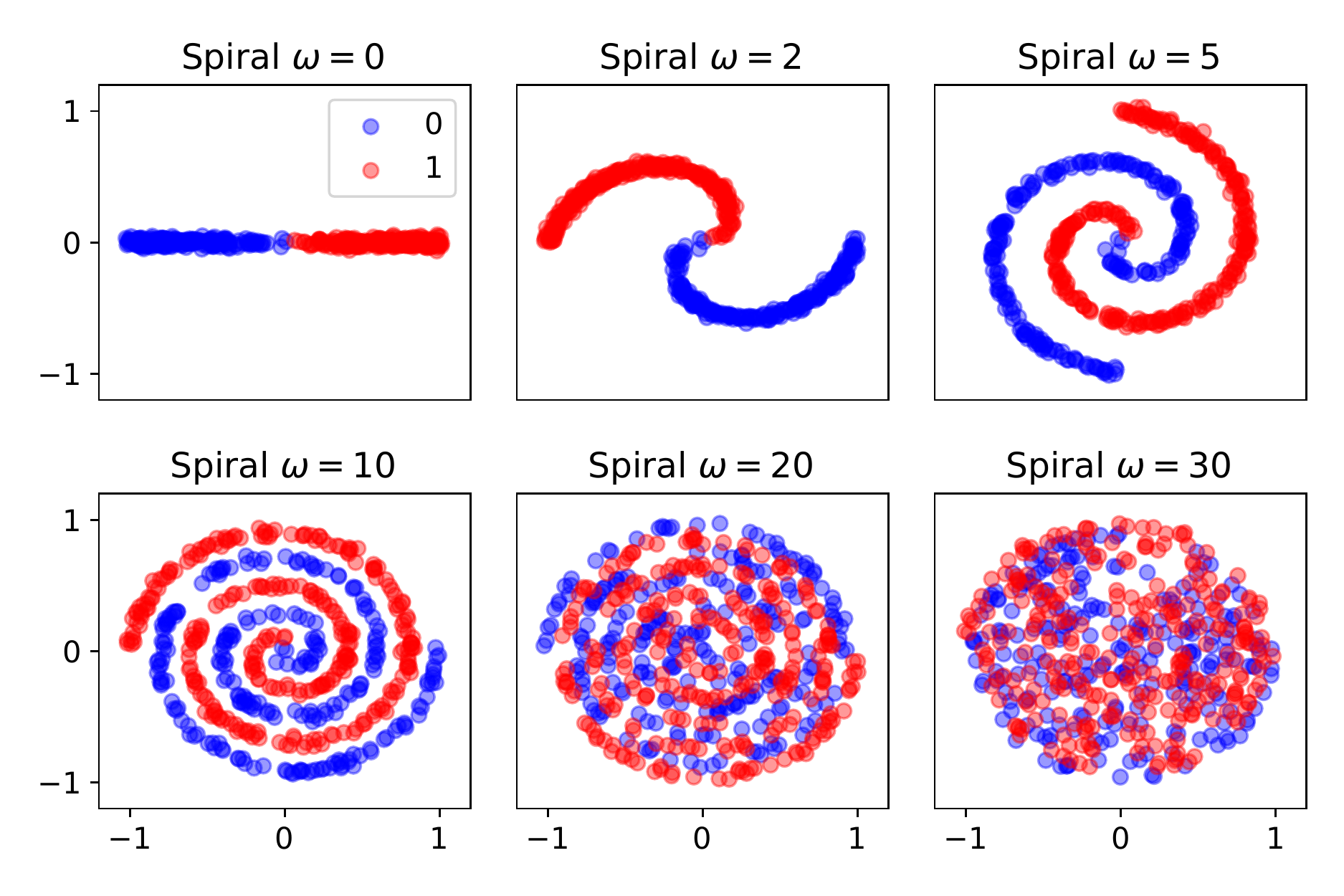}
    \caption{Spiral datasets $\mathcal{D}(\omega)$ for different rotation speed $\omega$. As $\omega$ increases, the two branches of the spiral become harder to distinguish. When $\omega$ reaches $30$, the data exhibits almost no pattern.}
    \label{fig:spirals}
\end{figure}
\parhead{Dataset.} To understand the role of depth, we highlight a natural question about neural network architecture: why do we need more depth when only a single wide enough hidden layer can approximate any smooth function \cite{hornik1989multilayer, barron1994approximation}? A half-answer would be that the same universal approximation theorem also holds with a deep enough neural network of fixed-width \cite{lu2017expressive}. A more complete answer lies in the quantification of the ``wide enough'' condition. Theoretical work on approximation theory has used small 2-hidden-layers neural networks to construct oscillating functions that cannot be approximated by any 1-hidden-layer network that uses a subexponential number of hidden units \cite{eldan2016power}. 

We adapt the construction of this paper and design labeled datasets $\mathcal{D}(\omega)$, each consisting of a spiral with two branches at rotation speed $\omega$. The binary labels correspond to each branch. When $\omega$ increases, the branches of the spiral get more interleaved and become harder to distinguish. Figure \ref{fig:spirals} shows examples and intuition. At $\omega = 30$, the organization of labels appears to be almost random. The mathematical details for the generation of the datasets are in appendix.

\begin{figure*}[t]
    \centering
    \begin{subfigure}[b]{0.49\textwidth}
         \centering
         \includegraphics[width=\linewidth]{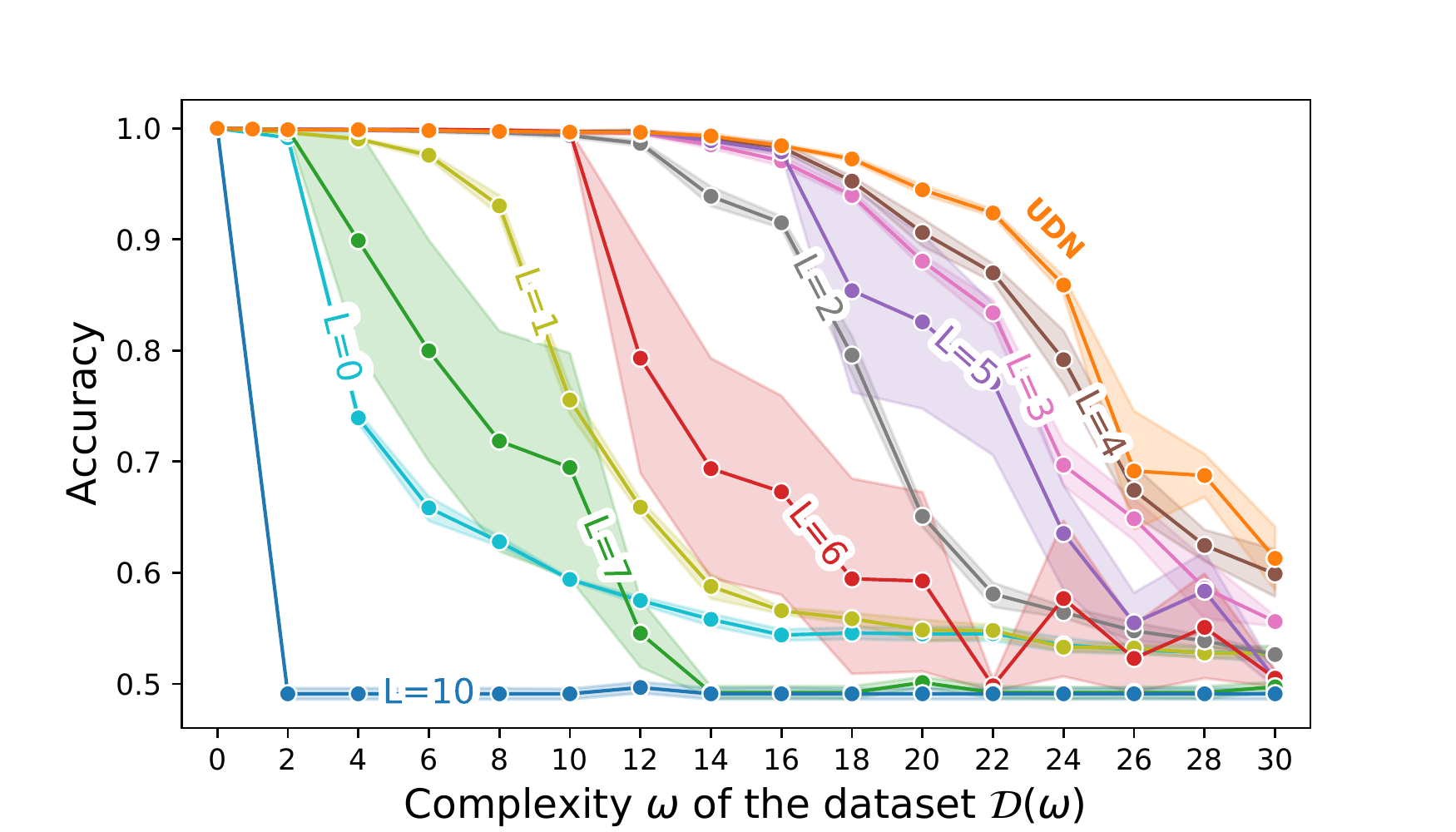}
         \caption{Classification accuracy for a range of models and datasets $\mathcal{D}(\omega)$. The UDN (orange) achieves the best accuracy for every dataset.}
         \label{fig:spirals-acc}
     \end{subfigure}
     \hfill
     \begin{subfigure}[b]{0.47\textwidth}
         \centering
         \includegraphics[width=\textwidth]{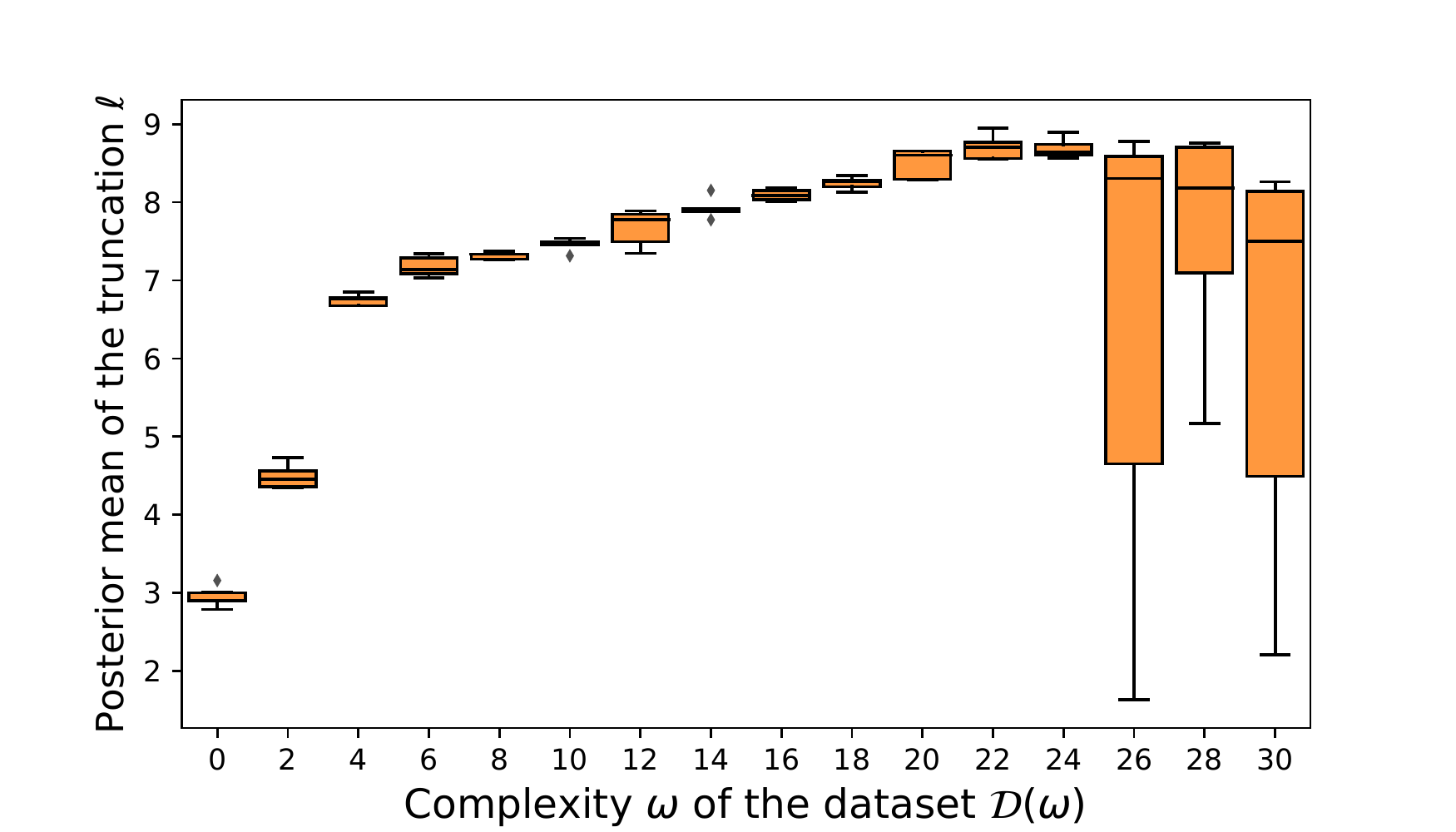}
         \caption{Mean depth $\ell$ of the UDN posterior, fitted for different $\mathcal{D}(\omega)$. The UDN becomes deeper when the complexity increases.
         }
         \label{fig:spiral-posterior}
     \end{subfigure}
    \caption{(a)  For all models, the accuracy decreases when $\omega$ increases. 
    The accuracy for all models drops at large $\omega$, which corresponds to when the datasets start to lose their patterns; see Figure \ref{fig:spirals}. (b) The UDN favors larger depths as $\omega$ increases. It adapts to the complexity of the data. Near $\omega \sim 30$, independent runs of the inference yield different posterior expected depths. It happens when the accuracy drops in (a), and has the same explanation: when the dataset contains almost random label patterns, the UDN does not have an optimal depth to explain the data.  -- The error bands and the boxes with whiskers are computed with 5 independent runs for each dataset and model.}
\end{figure*}

\parhead{Experiment.} We follow the notations of section \ref{sec:model}. The architecture generator $f$ is defined to return layers with $32$-hidden units each. So $f_1: \R^2 \rightarrow \R^{32}$ and $f_\ell: \R^{32} \rightarrow \R^{32}$ for $\ell \geq 2$. Each $f_l$ is a linear function followed by a ReLU activation. The output layers $o_\ell$ are linear layers transforming a hidden state of $\R^{32}$ to a vector in $\R^2$, whose \textit{softmax} parametrizes the probabilities of the two labels.

We train finite neural networks $\Omega_L$ for a large range of depths $L$, using the architecture returned by $f$. Each $\Omega_L$ and the UDN are trained independently. We also train Depth Uncertainty Networks \cite{antoran2020depth} with an ensemble of depths 1 to 10 -- referred to as DUN-10.

For each $\omega$, we generate a dataset $\mathcal{D}(\omega)$ on which we train the models for 4000 epochs. We then select the best epoch using a validation set and report the accuracy on a test set.

\parhead{Results.}
Figure \ref{fig:spirals-acc} reports the test accuracy of the UDN and the finite models for $\omega \in [0,30]$. The UDN achieves the best accuracy for every dataset complexity. 

Figure \ref{fig:spiral-posterior} offers reasons for the success of the UDN. The posterior UDN places mass on increasingly deeper $\ell$ when $\omega$ increases. Interestingly, at $\omega \geq 26$, the posterior of the UDN does not select a precise depth across independent runs. This behavior is coherent with the drop in accuracy in figure \ref{fig:spirals-acc} since the dataset contains label patterns that are almost random at large $\omega$, and the UDN does not have an optimal depth to explain the data.

Table \ref{tab:spiral-table} reports the average accuracy (across $\omega$) for the best models and DUN-10. The UDN outperforms all of the other models. DUN-10 also outperforms the individual $\Omega_L$, suggesting that combining several depths of the same network, like the posterior predictive of the UDN does, improves the representation capability of neural networks.

\begin{table}[h]
    \centering
    \begin{tabular}{lc}
    \toprule
        Model & Average accuracy ($\%$) \\
    \midrule
        Standard network $\Omega_5$ & $84.8 \pm 3.6$ \\
        Standard network $\Omega_3$ & $87.4 \pm 0.6$ \\
        Standard network $\Omega_4$ & $89.3 \pm 0.9$ \\
        DUN-10 \cite{antoran2020depth} & $90.5 \pm 0.7$ \\
        \textbf{UDN} & $\mathbf{91.7 \pm 1.1}$ \\
    \bottomrule
    \end{tabular}
    \caption{Test accuracy averaged over the different dataset $\mathcal{D}(\omega)$. The UDN outperforms all the other models. Standard deviations are calculated on 5 runs.}
    \label{tab:spiral-table}
\end{table}

\subsection{Image classification on CIFAR-10 with CNN}
\label{subsec:cifar10}
\parhead{Dataset.}
Image classification is a domain where deeper networks have pushed the state-of-the-art. We study the performance of the UDN on the CIFAR-10 dataset \cite{krizhevsky2009learning}. We use a layer architecture\footnote{We are interested in the inference of the UDN and not in improving the state-of-the-art of image classification. Hence, we did not tune the best architecture possible.} adapted from \citet{he2016resnet}. ResNet building blocks are a succession of three convolution layers with a residual shortcut from the input to the output of the block. Furthermore, Batchnorm \cite{ioffe2015batch} and ReLU follow each convolution. The exact dimensions of the convolutions are in appendix.

\parhead{Experiment.}
Using the notations of section \ref{sec:model}, we define the architecture generator $f$ to return layers $f_\ell$ that are ResNet blocks. Each block contains 3 convolutions, so each network $\Omega_L$ corresponds to a ResNet-$3L$. 
The output layers linearly map hidden states to vectors of $\mathbb{R}^{10}$, whose \textit{softmax} parametrize the probabilities of the ten image classes.

We evaluate the UDN and various $\Omega_L$. We report the performance of the ResNet-18 trained in \citet{bai2020multiscale}, and the performance of implicit methods: Neural ODEs (NODE), Augmented Neural ODEs (ANODE) and Multiscale Deep Equilibrium Models (MDEQ) \cite{chen2018neural, dupont2019augmented, bai2020multiscale} which are competitive approaches for infinitely deep neural networks.

To evaluate the UDN and the $\Omega_L$, we use the default train-test split of CIFAR-10, and report the test accuracy at the end of training. We use the same hyperparameters and architectures for the UDN and the individual $\Omega_L$. Following \citet{he2016resnet}, we use the SGD optimizer and a specific learning rate schedule, detailed in appendix. Table \ref{tab:cifar10} reports the results. The UDN outperforms all the other models. 
\begin{table}
    \centering
    \begin{tabular}{lcc}
    \toprule
        Model & Accuracy \\
    \midrule
        ResNet-15, $\Omega_5$ & $91.7 \scriptstyle \pm 0.2 $            \\ 
        ResNet-18 \cite{bai2020multiscale} & $92.9 \scriptstyle\pm 0.2$ \\ 
        ResNet-24, $\Omega_8$ & $93.6 \scriptstyle\pm 0.4 $             \\ 
        ResNet-30, $\Omega_{10}$ & $94.0 \scriptstyle\pm 0.2 $          \\ 
        ResNet-45, $\Omega_{15}$ & $94.0 \scriptstyle\pm 0.2 $          \\ 
        ResNet-60, $\Omega_{20}$ & $93.9 \scriptstyle\pm 0.1 $          \\ 
        ResNet-90, $\Omega_{30}$ & $93.9 \scriptstyle\pm 0.1 $          \\ 
    \midrule
        NODE \cite{dupont2019augmented} & $53.7 \scriptstyle\pm 0.2$    \\ 
        ANODE \cite{dupont2019augmented} & $60.6 \scriptstyle\pm 0.4$   \\ 
        MDEQ \cite{bai2020multiscale} & $93.8 \scriptstyle\pm 0.3$      \\ 
    \midrule
        \textbf{UDN with ResNet} & $\mathbf{94.4 \scriptstyle\pm 0.2}$       \\ 
    \bottomrule
    \end{tabular}
    \caption{Test accuracy on CIFAR-10. The UDN outperforms all the other models. With the architecture that we use, the best finite models seem to have $L$ between 10 and 20 (ResNet-30 and ResNet-60). -- Standard deviations are calculated on 3 runs for our experiments, 5 runs for the reported ones.}
    \label{tab:cifar10}
\end{table}

\parhead{Exploration of the posterior space.}
Finally, we aim to analyze how the posterior depth adapts to the complexity of the data. For this purpose, we create two subsamples of CIFAR-10: an \textit{easy} subsample containing only \textit{deer} and \textit{car} images, and a \textit{hard} subsample containing only \textit{cat} and \textit{dog} images. These categories were selected from an independently generated CIFAR-10 confusion matrix, in which (deer,car) were found to be the least confused labels, whereas (cat,dog) were the most. We suspect that the \textit{hard} dataset will require more layers than the \textit{easy} dataset, and will cause the posterior of the UDN to adapt accordingly.

\begin{figure}
    \centering
    \includegraphics[width=\linewidth]{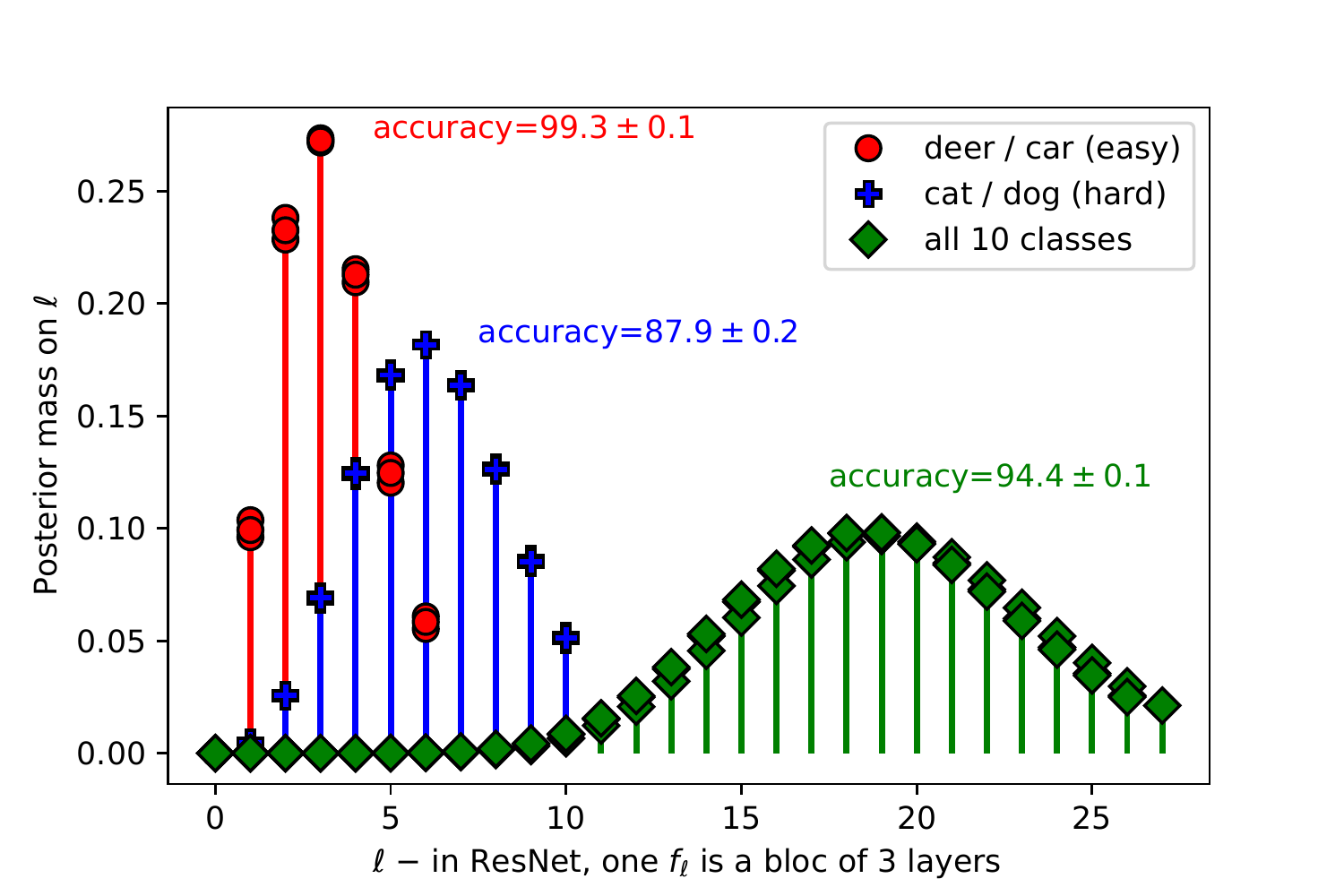}
    \caption{The UDN posterior on the depth $\ell$ adapts to the complexity of the dataset. It uses deeper truncations for a harder classification task like \textit{cat vs dog} than for an easier one like \textit{deer vs car}. When classifying the full CIFAR-10, the posterior UDN covers an even deeper set of truncations, which matches the depths of the best finite models from table \ref{tab:cifar10}. The maximal truncation considered by the green (diamond) posteriors corresponds to a ResNet-81. -- For each dataset, 3 independent runs  are represented.}
    \label{fig:posterior-depth}
\end{figure}

We infer the posterior of the UDN on each of these two subsampled datasets, in addition to the full CIFAR-10. Figure \ref{fig:posterior-depth} reports the probability mass functions of the posteriors. For the easiest pair of labels (deer, car), the UDN only uses a couple of layers to reach $99\%$ accuracy, whereas it uses more layers for the harder pair (cat,dog) but reaches only $88\%$. For the full dataset, the posterior puts mass on much more layers and the UDN achieves $94\%$ accuracy. The posteriors effectively adapt to the complexity of the datasets. 

\section{Discussion and future work}
We proposed the unbounded depth neural network, a Bayesian deep neural network that uses as many layers as needed by the data, without any upper limit. We demonstrated empirically that the model adapts to different data complexities and competes with finite and infinite models. 

To perform approximate inference of the UDN, we designed a novel variational inference algorithm capable of managing an infinite set of latent variables. We showed with experiments on real data that the algorithm successfully explores different regions of the posterior space. 

The UDN and the dynamic variational inference offer several avenues for further research. First, the unbounded neural network could be applied to transformers, where very deep models have shown successful results \cite{liu2020deepnmt}. Another interesting direction is to use the unbounded variational family for variational inference of other infinite models, such as Dirichlet processes mixtures. Reciprocally, the UDN could be extended into a deep nonparametric model, where the truncation $\ell$ changes across observations in the same dataset. 
Finally, future studies of the UDN could examine how the form of the neural network's weight priors impacts the corresponding posteriors.




\section*{Acknowledgments}
We are thankful to Gemma Moran, Elham Azizi, Clayton Sanford and anonymous reviewers for helpful comments and discussions. This work is funded by NSF IIS 2127869, ONR N00014-17-1-2131, ONR N00014-15-1-2209, the Simons Foundation, the Sloan Foundation, and Open Philanthropy.
\bibliography{main}
\bibliographystyle{icml2022}

\newpage
\appendix
\onecolumn
\section{The truncated Poisson family}
We prove Theorem \ref{th:tp}:
For any $\delta \in [0.5, 1[$, $\mathcal{TP}(\delta)$ is unbounded with connected bounded members. 
For $\delta=0.95$, we have

\begin{enumerate}
    \item $\lambda - \ln 2 \leq m(q^{{0.95}}(\lambda)) \leq 1.3 \lambda + 5$ \hspace{0.93cm} (\ref{tp:upper})
    \item $\forall n\in \N, ~~ n \in \textnormal{argmax}(q^{0.95}(n+0.5))$  \hspace{0.4cm}(\ref{tp:argmax})
    \item $\lambda > 0$ ~is a continuous parameter. \hspace{1.16cm} (\ref{tp:continuous})
\end{enumerate}

\begin{proof}
We first show that $\mathcal{TP}(\delta)$ is unbounded with connected bounded members for any $\delta \in [0.5, 1[$.

\begin{itemize}
    \item Each distribution is the truncation of a Poisson distribution so it has a finite support by definition. Hence, $\mathcal{TP}(\delta)$ has bounded members and satisfies (\ref{cond:bounded}).
    \item The assertion (\ref{tp:argmax}) is true for any $\delta$, we show it below. It ensures that $\mathcal{TP}(\delta)$ is unbounded and satisfies (\ref{cond:mode}).
    \item $\lambda$ is a continuous positive real number by design of the family. Hence the members of $\mathcal{TP}(\delta)$ are continuously connected and the family satisfies (\ref{cond:continuous}).
\end{itemize}
So $\mathcal{TP}(\delta)$ is unbounded with connected bounded members for any $\delta \in [0.5, 1[$.\\

We prove the additional assertions.
By definition of the quantile truncation, we will use that $m(q^\delta(\lambda)) \geq m(q^\eta(\lambda))$ if $\delta \geq \eta$, and that $m(q^\delta) \leq \delta$-quantile of $q$.

\begin{itemize}
    \item[1.] \textbf{Lower bound}: According to \cite{choi1994medians}, the median of $\text{Poisson}(\lambda)$ is greater or equal than $ \lambda - \ln 2$. Hence, by definition of the quantile truncation, we have $m(q^\delta(\lambda)) \geq \text{median}(\text{Poisson}(\lambda)) \geq \lambda - \ln 2$ for any $\delta \geq 0.5$.
    \item[2.] The modes of a Poisson($\lambda$) are $\ceil{\lambda}-1$ and $\lfloor \lambda \rfloor$ (which are the same value for non-integer $\lambda$). For a truncation at level $\delta$ with $\delta \geq 0.5$, the lowest term truncated can at most reach the median. Since $\text{median}(\text{Poisson}(\lambda)) \geq  \lambda - \ln 2$, the lowest truncated term of $\text{Poisson}(n+0.5)$ is at most $n + 0.5 - \log 2 > n$. But we also have $n =\lfloor n+0.5 \rfloor$. So the mode $n$ does not get truncated. Since the mode is not affected by re-normalization of the probabilities, $n$ is still the mode of $\text{Poisson}(n+0.5)$.
    \item[1.] \textbf{Upper bound}: Let $X \sim \text{Poisson}(\lambda)$ for some $\lambda > 0$.
    \begin{itemize}
        \item We recall a standard Chernoff bounds argument from \cite{mitzenmacher2017probability}\footnote{And also Wikipedia \url{https://en.wikipedia.org/wiki/Poisson_distribution\#Other_properties}}:\\ $\forall x > \lambda, \mathbb{P}(X \geq x) \leq \frac{(e\lambda)^x e^{-\lambda}}{x^x}$.
        \item For $x = 1.3 \lambda$, it yields: 
        $\mathbb{P}(X \geq 1.3\lambda) \leq e^{0.3 \lambda} \parens{\frac{1}{1.3}}^{1.3 \lambda} = \parens{e^{0.3} \frac{1}{1.3^{1.3}}}^\lambda < 0.96^\lambda$. \\
        For $\lambda > 70$, we have $ \mathbb{P}(X \geq 1.3\lambda) \leq 0.96^{70} < 0.05$, so $m(q^{0.95}(\lambda)) \leq 1.3\lambda \leq 1.3\lambda + 5 $. 
        \item Note that for $\mu \geq \lambda$, we have $m(q(\mu)) \geq m(q(\lambda))$. In particular, we have: $m(q(\lambda)) \leq m(q(\ceil{\lambda}))$. Also we have $1.3 \lfloor \lambda \rfloor + 5 \leq 1.3 \lambda + 5$. So it suffices to show that $\forall \lambda \in ]0, 70], m(q(\ceil{\lambda})) \leq 1.3 \lfloor \lambda \rfloor + 5$ and we will have that $\forall \lambda \in ]0, 70], m(q(\lambda)) \leq 1.3 \lambda + 5$.\\
        It suffices to check manually that, $\forall k \in \curly{1,2,...,70}, m(q(k)) \leq 1.3 (k-1) + 5$. We check this in Python and report the figure \ref{fig:truncated-poisson-check} where each $(1.3 (k-1) + 5) - m(q(k))$ is positive for $k \in \curly{1,2,...,70}$.
        \begin{figure}[h]
            \centering
            \includegraphics[scale=0.5]{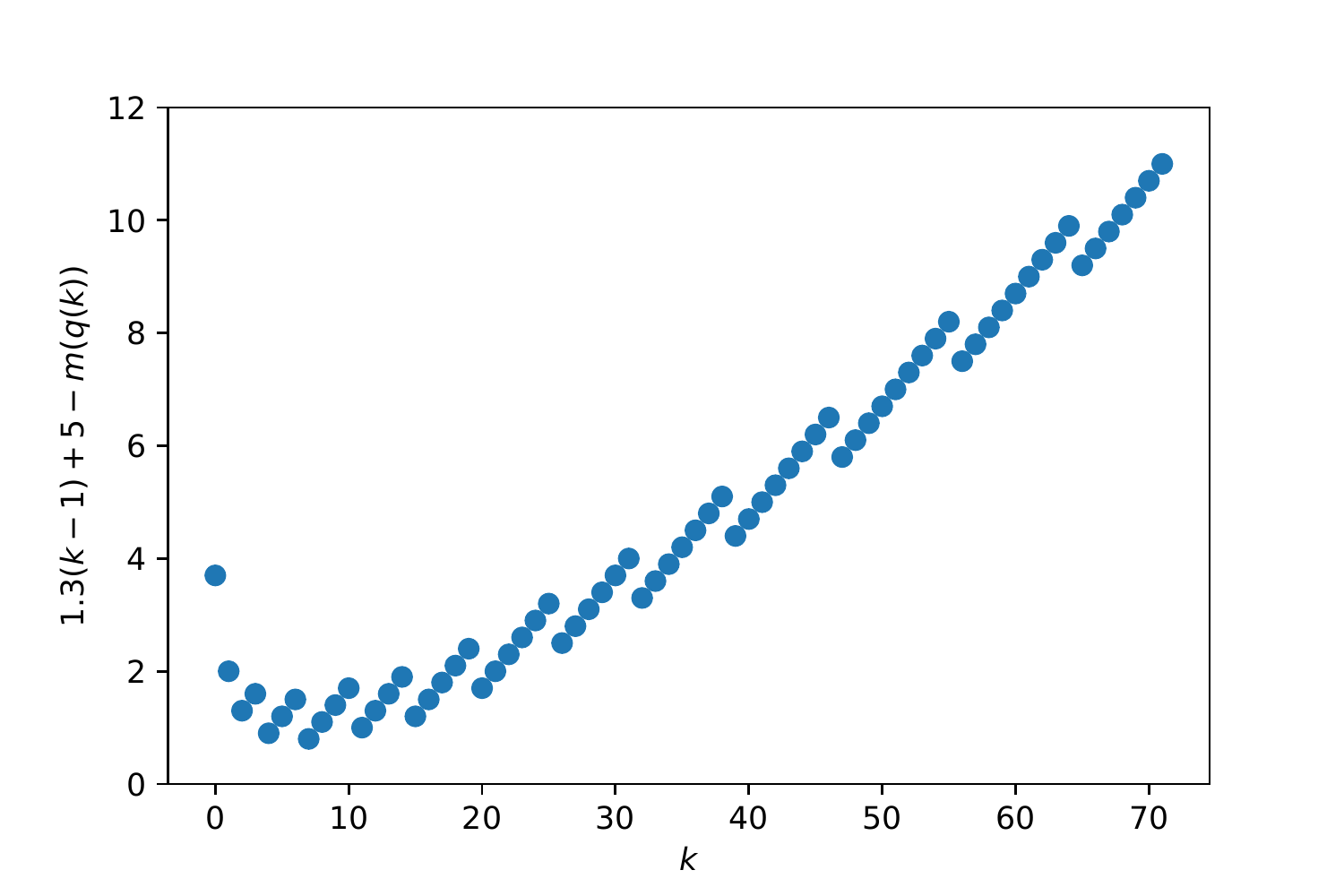}
            \caption{Empirical evaluation of $\delta_k = (1.3 (k-1) + 5) - m(q(k))$ for $k \in \curly{1,2,...,70}$. We observe that $\delta_k \geq 0$. }
            \label{fig:truncated-poisson-check}
        \end{figure}
        \item This concludes the proof.
    \end{itemize}
    \item This concludes the proof for the three assertions about $\delta=0.95$.
\end{itemize}
\end{proof}

\section{Experiments}

\subsection{Synthetic experiment}
\parhead{Dataset.}
The spiral dataset $D(\omega) = \curly{(x_i,y_i)}$ is generated by the following generative process:
\begin{align}
    t &\sim \text{Uniform}([0,1])\\
    u &= \sqrt{t} \\ 
    y &\sim \text{Uniform}(\curly{-1,1}) \\
    x &\sim \mathcal{N}\parens{\bmat{yu \cdot \cos\parens{\omega u \cdot \frac{\pi}{2}}  \\ yu \cdot \sin\parens{\omega u \cdot \frac{\pi}{2}}}, 0.02}
\end{align}
The square root of $t$ is taken to rebalance the density along the curve $u \mapsto \bmat{u \cdot \cos\parens{\omega |u| \cdot \frac{\pi}{2}} \\ u \cdot \sin\parens{\omega |u| \cdot \frac{\pi}{2}}}$.

\begin{figure}[t]
    \centering
    \begin{subfigure}[b]{0.48\textwidth}
         \centering
         \includegraphics[width=\linewidth]{fig/spiral-acc.pdf}
         \caption{Learning rate $0.005$}
         \label{fig:spirals-acc-005}
     \end{subfigure}
     \hfill
     \begin{subfigure}[b]{0.48\textwidth}
         \centering
         \includegraphics[width=\textwidth]{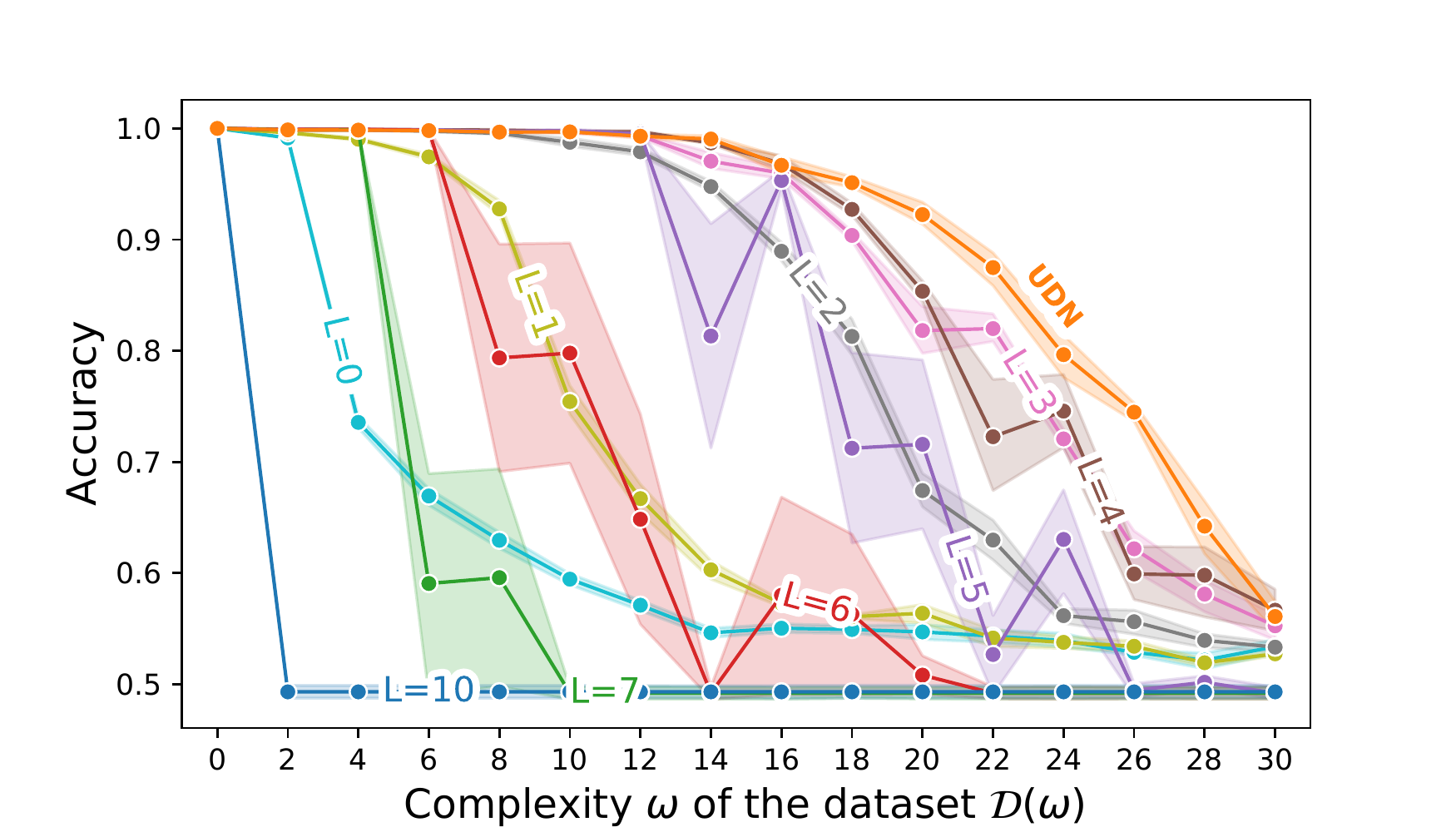}
         \caption{Learning rate $0.01$
         }
         \label{fig:spirals-acc-010}
     \end{subfigure}
    \caption{Classification accuracy for a range of models and datasets $\mathcal{D}(\omega)$. The
UDN (orange) achieves the best accuracy for almost each complexity $\omega$. The deep finite models ($L\geq 5$) are sensitive to a higher learning rate, meanwhile the UDN is robust even with its large (infinite) depth. -- The error bands are computed with 5 independent runs for each pair of dataset and model.}
\label{fig:compare-lr-spiral}
\end{figure}

\parhead{Training.}
For each $\omega$, we independently sample a train, a validation and a test dataset of each $1024$ samples.
We use the following hyperparameters for training:
\begin{itemize}
    \item Prior on the neural network weights: $\theta \sim \mathcal{N}(0,1)$
    \item Prior on the truncation $\ell$: $\ell - 1 \sim \text{Poisson}(0.5)$. We recall that $\ell \geq 1$ so we shift the Poisson by $1$.
    \item Optimizer: Adam \cite{kingma2014adam}
    \item Learning rate: $0.005$. Results with different learning rates are compared in Figure \ref{fig:compare-lr-spiral}.
    \item Learning rate for $\lambda$: we use a learning rate  that is $1/10$th of the general learning rate of the neural network weights.
    \item Initialization of the variational truncated Poisson family: $\lambda = 1.0$
    \item Number of epochs: 4000
\end{itemize}

\subsection{CIFAR-10 experiment}
\parhead{Architecture generator $f$}
We detail the construction of the architecture generator $f$ for the CIFAR experiment of section \ref{sec:experiments}.
A ResNet building block $B_k$ is the succession of three convolution layers $[1 \times 1 \times 2^k]$, $[3 \times 3 \times 2^k]$, and $[1 \times 1 \times 2^{k+2}]$ with a residual shortcut from the input to the output of the block. In these blocks, Batchnorm and ReLU follow every convolution. The architecture generator used by the UDN and the finite neural networks is the following:
$$ f: \ell \in \N \mapsto f_l := \left\{ \begin{array}{ll}
     B_6 & \text{if } \ell \in \bbrackets{1,3}\\
     B_7 & \text{if } \ell \in \bbrackets{4,8} \\
     B_8 & \text{else } 
\end{array} \right.$$

\parhead{Output generator $o$}
The output generator $o_\ell$ at each depth $\ell$ receives the hidden state $h_\ell$, performs a 2D average pooling with kernel $(4,4)$ and applies a linear layer of the adequate dimension to $\mathbb{R}^{10}$.

\subsection{Hyperparameters}

\begin{figure}
    \centering
    \includegraphics[scale=0.6]{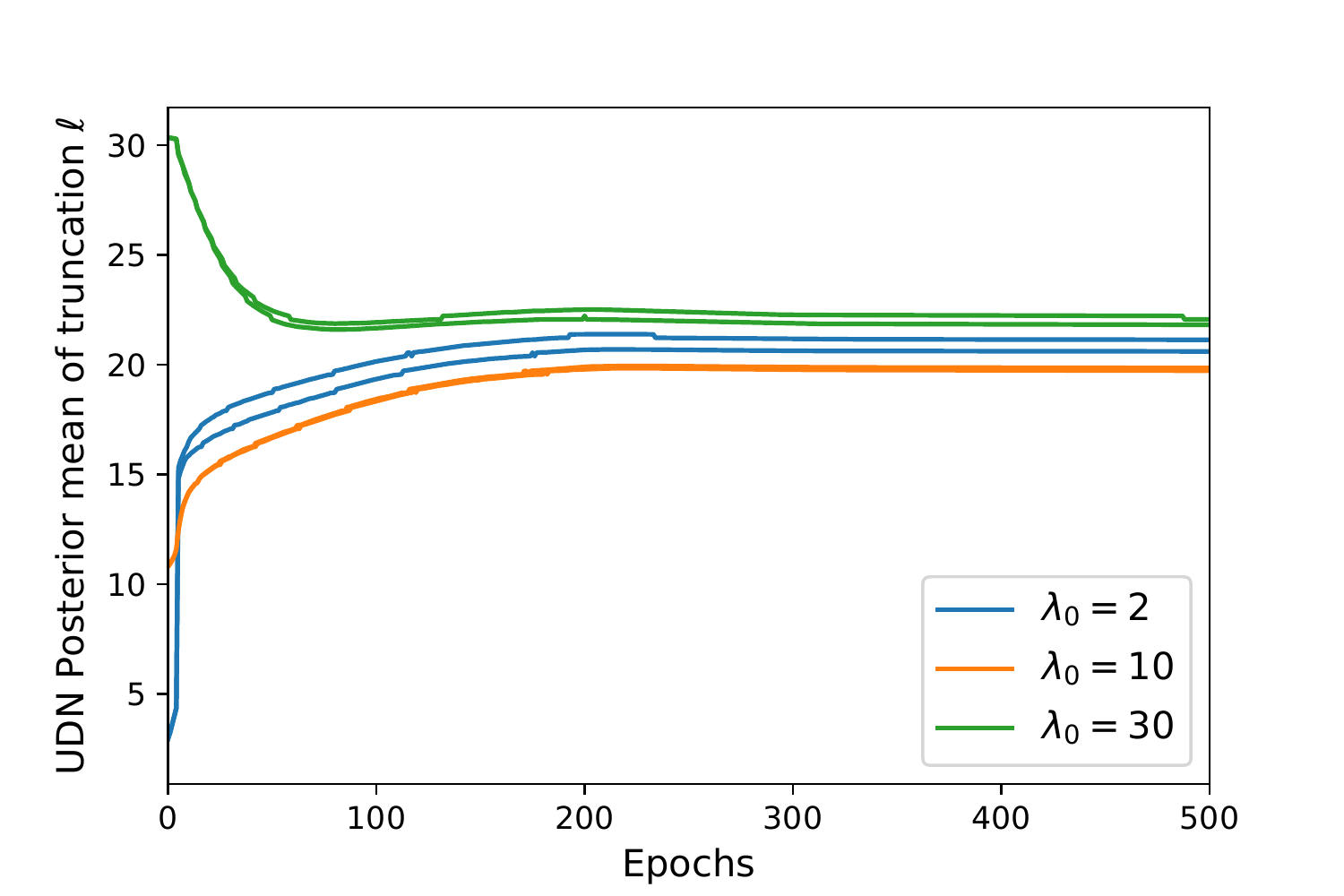}
    \caption{Evolution of the posterior mean of the truncation over multiple training with different initializations of the variational family. Trained on CIFAR-10. In all cases, the variational families explore the posterior space and converge to similar regions.}
    \label{fig:evolution}
\end{figure}

\parhead{Training.}
For the training on CIFAR-10, we follow the optimizer and the learning rate schedule from \citet{he2016resnet}. We increase our prior on the truncation since we suspect that CIFAR requires deep architectures. Similarly, we initialize the variational truncated Poisson family to $\lambda_0=5$. We show the trajectories of the variational families during the optimization for different values of $\lambda_0$ in figure \ref{fig:evolution}.
\begin{itemize}
    \item Prior on the neural network weights: $\theta \sim \mathcal{N}(0,1)$
    \item Prior on the truncation $\ell$: $\ell - 1 \sim \text{Poisson}(1)$.
    \item Optimizer: SGD with \texttt{momentum = 0.9, weight\_decay=1e-4}
    \item Number of epochs: 500
    \item Learning rate schedule: \texttt{[0.01]*5 + [0.1]*195 + [0.01]*100 + [0.001]*100}
    \item Learning rate for $\lambda$: we used the same learning rate for $\lambda$ and the weights
    \item Initialization of the variational truncated Poisson family: $\lambda_0 = 5.0$.
\end{itemize}

\subsection{Additional experiments on Regression datasets}
\begin{figure}
    \centering
    \includegraphics[scale=0.6]{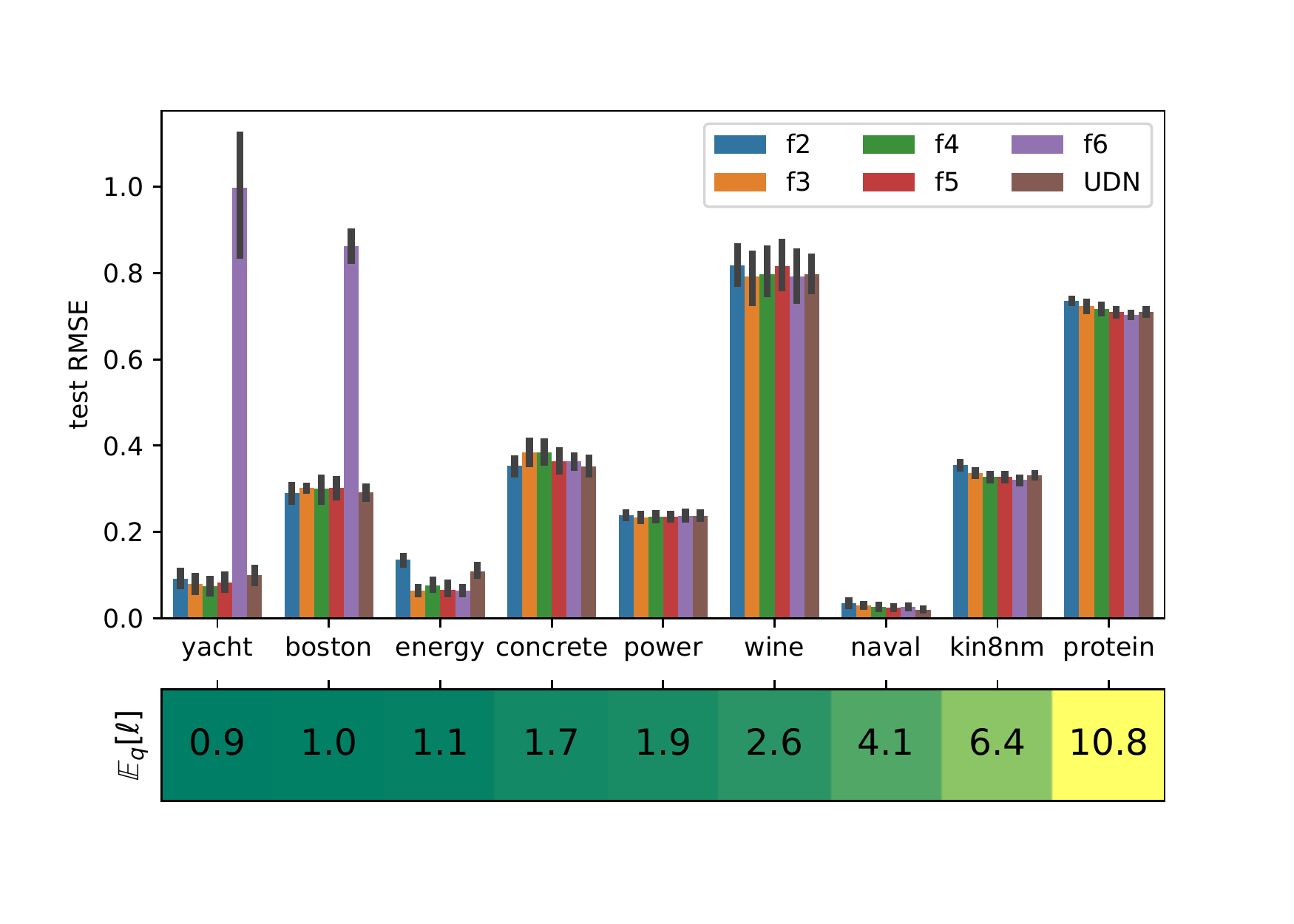}
    \caption{(top) Regression root mean square error (RMSE) on the test set of 9 UCI datasets. The optimal training epoch is chosen with a validation set. Error bars are obtained by repeating the experiment on 10 different train-valid-test sets. (bottom) Expectation of the number of layers $\ell$ in the UDN posterior. For the first five datasets, the expectation is lower than two. For the last four datasets, the number of layers gets higher and the UDN shows competitive performances against the finite alternatives.}
    \label{fig:regression}
\end{figure}

We run additional experiments on tabular datasets. We perform regression with the UDN for nine regression datasets from the UCI repository \cite{Dua:2019}:
\textit{Boston Housing} (boston) 
\textit{Concrete Strength} (concrete),
\textit{Energy Efficiency} (energy),
\textit{Kin8nm} (kin8nm),
\textit{Naval Propulsion} (naval),
\textit{Power Plant} (power),
\textit{Protein Structure} (protein),
\textit{Wine Quality} (wine) and
\textit{Yacht Hydrodynamics} (yacht).
For regression (instead of classification in the previous experiments), the UDN models the response variable with a Gaussian distribution whose mean is given by the output layers of the infinite neural network. In figure \ref{fig:regression}, we show the performance of the UDN against the finite variants (f2, f3, f4, f5, f6 with respectively exactly 2, 3, 4, 5, and 6 layers). Figure \ref{fig:regression} also indicates the posterior mean of the truncation level $\ell$ learned during inference. Overall, we notice that for regression tasks on tabular data, very deep networks are not necessary to achieve good prediction performance. For the datasets yacht, boston, energy, concrete and power, the posterior $\ell$ is lower than two and the UDN offers similar performances than the finite alternatives. For the other datasets such as wine, naval, kin8nm, and protein, the UDN learns a higher number of layers and performs competitively or better than the finite alternatives. 

\end{document}